# EXPERT PC TROUBLESHOOTER WITH FUZZY-LOGIC AND SELF-LEARNING SUPPORT

Youssef Bassil

LACSC – Lebanese Association for Computational Sciences
Registered under No. 957, 2011, Beirut, Lebanon

youssef.bassil@lacsc.org

## ABSTRACT

*Expert systems use human knowledge often stored as rules within the computer to solve problems that generally would entail human intelligence. Today, with information systems turning out to be more pervasive and with the myriad advances in information technologies, automating computer fault diagnosis is becoming so fundamental that soon every enterprise has to endorse it. This paper proposes an expert system called Expert PC Troubleshooter for diagnosing computer problems. The system is composed of a user interface, a rule-base, an inference engine, and an expert interface. Additionally, the system features a fuzzy-logic module to troubleshoot POST beep errors, and an intelligent agent that assists in the knowledge acquisition process. The proposed system is meant to automate the maintenance, repair, and operations (MRO) process, and free-up human technicians from manually performing routine, laborious, and time-consuming maintenance tasks. As future work, the proposed system is to be parallelized so as to boost its performance and speed-up its various operations.*

## 1. INTRODUCTION

An expert system sometimes referred to as knowledge-based system is a computer software that emulates the decision-making ability of a human expert [1]. In essence, expert systems do not use traditional programming paradigms to solve problems; rather, they use knowledge which they reason about to draw conclusions and provide solutions. Expert system is a subfield of artificial intelligence (AI), and was first conceived by Edward Feigenbaum, now considered the father of expert systems, who with other colleagues and associates built the first successful expert system in the late 60's at Stanford University [2]. It was called the "Dendral" system, a portmanteau of the term "Dendritic Algorithm". Dendral was meant to emulate organic chemists to help automating the process of identifying unknown organic molecules [3, 4]. The advantage of expert systems over conventional programs is that their core algorithm is not encapsulated in the programming code but stored as knowledge in an independent database called knowledge-base or KB. In consequence, there is no need for the expert system to be reprogrammed and recompiled every time the knowledge changes. Practically, expert systems have significant applications including medical diagnosis, fault diagnosis, question-answering, industrial process controlling, climate forecasting, manufacturing failure analysis, decision support, and decision making [5]. Today, with the prevalent use of computers, fault diagnosis is becoming crucial in the field of computer engineering and information technology, particularly in personal computer troubleshooting; however, acquiring the troubleshooting knowledge from expert computer technicians is limited as it requires continuous learning, training, and practice in maintenance skills which on the long run can dramatically increase organization operating costs, decrease their net productivity, and proliferate their revenue leakage and losses. Basically, PC (IBM Personal Computer) troubleshooting covers a wide spectrum of problems including hardware problems, software problems, network problems, server problems, operating system problems, and application software problems. This would impose a maintenance nightmare for large-scale enterprises and IT infrastructures as the number of possible technical problems and finding their solutions can become very huge and complex.

This paper proposes an expert system for troubleshooting PC problems and diagnosing computer hardware, software, and network faults. The system is a rule-based expert system called Expert PC Troubleshooter that can be deployed in a web Intranet environment, allowing different distributed users and machines to access the system over a network architecture. The actual knowledge of the system is represented as production rules in the form of *if <condition1> and <condition2> then <conclusion>* → *<solution>*, and stored in a rule-base. Characteristically, the proposed Expert PC Troubleshooter comprises six major modules: a GUI web user interface which allows human troubleshooters to easily operate and interact with the system; a knowledge-base more particularly a rule-base which houses and stores all PC troubleshooting knowledge as human-readable production rules; an inference engine which matches facts provided by human troubleshooters against rules in the rule-base, then produces a reasoning on these rules based on forward-chaining algorithm to derive conclusions and identify computer faults; a fuzzy-logic inference engine coupled with a fuzzy-logic rule-base which allows the diagnosis of BIOS POST beep errors that contain notions like very short beep, long beep, and very long beep; an autonomous intelligent agent for supporting knowledge acquisition and self-learning processes which scraps online troubleshooting knowledge located in predefined web resources and transforms them into production rules storable in the system's rule-base; and an admin portal or expert interface which facilitates adding, editing, updating, and deleting the system's production rules.

The proposed Expert PC Troubleshooter aids in decision-making and allows automated fault detection and deep problem diagnosis by emulating the human reasoning activity. From a business perspective, the system would reduce maintenance costs, provide quicker problem resolution, reduce employee training time, and deliver higher job quality.

## 2. RELATED WORK

Fault diagnosis in the engineering field has received a lot of attention in research literature over the last years. Several theoretical and practical fault troubleshooting techniques have been developed and experimented to automate the diagnostic process of electronic devices. "Dendral" was maybe the first successful expert system built in the 1965 at Stanford University by Edward Feigenbaum, Bruce Buchanan, Joshua Lederberg, and Carl Djerassi, along with other colleagues, associates, and students. The "Dendral" term was derived from "Dendritic Algorithm" and was aimed at helping organic chemists to analyze the mass spectrometry data of chemical substances to determine the structure of their organic molecules and identify if they are considered unknown molecules. This was in the search for evidence of extraterrestrial life [4, 6]. SpotLight [7] is yet another expert system based on case knowledge whose purpose is to troubleshoot unplanned maintenance tasks in the aerospace industry. Its prime objective is to overcome system unreliability and reduce the frequency of equipment failure, in addition to support unplanned events that are unanticipated by the manufacturers such as weather related anomalies. The system uses case knowledge to solve new problems by using already stored knowledge about similar problems. In a different context, a methodology for building an expert system for car troubleshooting and maintenance was proposed [8]. It simulates a human car mechanic to help car's owners diagnose their car problems and get suggestions on how to solve them. The system employs various set of rules to handle the different type of failures but more specifically starting problems and their sub-problems. In the same domain, architecture for building an expert system that diagnoses faults in electronic hydraulic braking systems was proposed [9]. The architecture combines the acquired knowledge with fuzzy knowledge and uses them for the final diagnostic reasoning. Knowledge generation and reasoning are handled by a software simulation. [10] proposed a knowledge-based system for troubleshooting complex equipment based on case-based reasoning (CBR). The system harnesses stored cases representing experience to support decision makers in solving new problems. The expert troubleshooter recognizes the root cause of the new problem based on the most similar case, and then selects the best matching case to assert the root cause and consequently to formulate the repair plan. [11] proposed an expert system for the troubleshooting and diagnosis of electronic devices. The knowledge is acquired during the development phase from computer aided design, in addition to engineering data provided by the design engineer. A supplementary interview session is conducted to get any other remaining knowledge. In the troubleshooting phase, when a new problem is presented, the system reasons on the knowledge already stored in the system so that the cause error is diagnosed and reported. Another popular system is "PCDIASHOOT" (PC DIAgnosis and troubleSHOOTing) [12] which was developed to assist PC technicians and computer users in troubleshooting their computers in case they fail to boot. The system supports the diagnosis of IBM compatible PCs during the power-on self-test (POST). POST is a routine that checks computer's hardware at boot time to ensure that they are present and functioning properly. "PCDIASHOOT" uses the POST generated troubleshooting codes to provide the troubleshooter with much more details on the fault a malfunctioning PC is having.

## 3. THE PROPOSED EXPERT PC TROUBLESHOOTER

The proposed Expert PC Troubleshooter is a rule-based expert system for diagnosing and troubleshooting PC faults and hardware problems. It is a web-based system that can deployed in an Intranet environment or hosted on the Internet. Typically, it is composed of six modules: a GUI user interface which eases the human-machine interaction; an knowledge-base or rule-base which stores the production rules of the system representing the troubleshooting knowledge; an inference engine which reasons on stored knowledge and user's facts to draw conclusions and provide solutions for identified faults; a fuzzy-logic inference engine together with a fuzzy-logic rule-base which diagnoses BIOS POST beep errors; an intelligent agent which assists in the self-learning and knowledge acquisition processes; and an expert interface which simplifies the manipulation and management of production rules by system's administrators. Figure 1 depicts the core architecture of the proposed Expert PC Troubleshooter system.

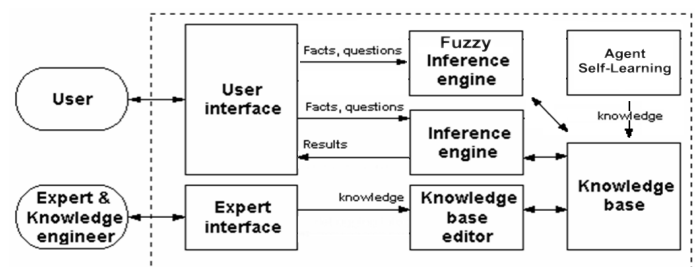

Figure 1. Expert PC Troubleshooter

## 3.1. The User Interface

The user interface of the proposed Expert PC Troubleshooter is a web-based interface accessible from any compatible web browser and allows a bi-directional communication between the system and the user. It is a sort of a step-by-step multi-page diagnostic questionnaire containing a sequence of questions that are asked to the user who would have to answer them thoroughly, so that the cause of the fault is identified and a matching solution is provided. In other words, it is web data form through which the possible symptoms of hardware faults are presented to the user on the screen. The user has then to select the symptoms that are exhibited by the malfunctioning PC so as to start a reasoning process to find the cause of the fault and its solution. Figure 2 shows a sample multiple-answer diagnostic question web-form asking the user for the type of problem he is having.

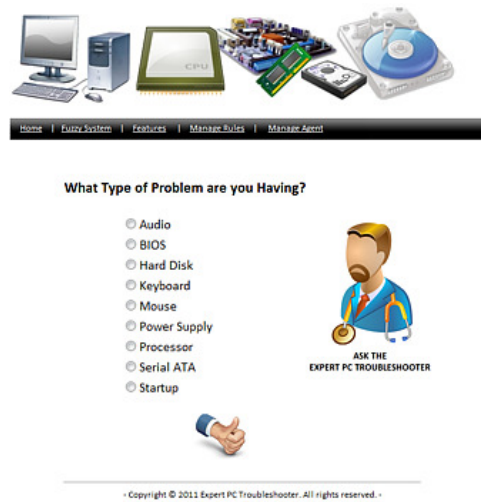

Figure 2. Expert PC Troubleshooter interface

## 3.2. The Knowledge-Base

The knowledge-base is a human-readable rule-base in which troubleshooting knowledge is represented as production rules originally acquired from human experts in the PC troubleshooting field. It is a data repository which provides a mean for knowledge to be collected, organized, saved, and searched. Upon reasoning, the expert system loads rules from the rule-base to the working memory and tries to match them against user's facts submitted via the user interface. Once a match occurs, the fault is identified and a solution is provided to fix that fault. Fundamentally, the rules of the proposed Expert PC Troubleshooter are in the form of *if-then* statements, more formally they can be represented as *IF A AND B THEN C → D* where *A* denotes the first condition, *B* denotes the second condition, *C* denotes the conclusion, and *D* denotes the solution. For instance, a possible production rule could be stated as follows: *IF hard disk problem AND SMART warning displayed THEN serious mechanical problems are detected → backup & replace your drive*. Table 1 delineates a subset of the production rules stored in the rule-base.

International Journal of Artificial Intelligence & Applications (IJAIA), Vol.3, No.2, March 2012

Table 1. Production rules.

| IF | AND | THEN | Solution |
|---|---|---|---|
| Audio | Sound Card can't be Detected. | Damaged or Sound Card Not Installed | Replace Sound Card |
| Audio | Driver Warning | Driver Conflict or Incompatible Driver | Install The Appropriate Driver |
| Audio | Scratchy Sound | Signal Interference | Stay Away from Radio Frequency Sources |
| Audio | Speaker or Microphone won't Work | Incorrect Jacks | Use Proper Jacks |
| BIOS | Calendar-Related and Leap-Year Bugs. | BIOS is out-of-date | Upgrade Flash BIOS |
| BIOS | Can't Install Flash BIOS Update | BIOS is Write-Protected | Disable Write-Protection |
| BIOS | Can't Access BIOS | BIOS is Password Protected | Remove Password |
| Hard Disk | Can't Access Full Capacity of Hard Drive over 8.4GB | BIOS is Out-of-Date | Upgrade BIOS |
| Hard Disk | Can't Use UDMA Drives at Full Speed. | BIOS is Out-of-Date or Incompatible IDE Cable | Upgrade BIOS or Replace IDE Cable |
| Hard Disk | IDE Drive not Ready Errors During Startup | Drive not Spinning Up Fast Enough at Startup | Enable or Increase Hard Disk Predelay-time |
| Hard Disk | Invalid Drive Specification Error | Drive has not Been Partitioned | Run FDISK to Create Valid Partitions |
| Hard Disk | Invalid Media Type Error | Drive not Yet Formatted | Format your Drive |
| Hard Disk | SMART Warning Displayed | Serious Mechanical Problems are Detected | Backup & Replace Your Drive |
| Keyboard | Num Lock Stays Off when Starting System. | Num Lock Shut off in BIOS | Turn on Num Lock in BIOS |
| Keyboard | Intermittent Keyboard Failures | Keyboard Cable or Keyboard Jack Might be Defective | Test Keyboard Cable or Jack with Digital Multimeter |
| Keyboard | Keys are Sticking | Keyboard might have Spilled Drink or Trapped Debris under Keys | Remove Keytops and Clean under Keys, or Wash-out Keyboard |
| Mouse | Mouse can't be Detected | Hardware Resource Conflict | Use Windows Device Manager to Find Conflicts and Resolve them |
| Mouse | Can't use PS/2 Mouse | PS/2 Mouse Port Might be Disabled | Enable PS/2 Mouse Port |
| Mouse | Mouse Pointer Jerks Onscreen | Mouse Ball or Rollers are Dirty | Clean Mouse Mechanism |
| Mouse | Mouse Works in Windows, but Not When Booted to DOS | DOS Driver Must be Loaded from AUTOEXEC.BAT or CONFIG.SYS | Install DOS Mouse Driver, and Reference it in Startup Files |
| Power Supply | System Reboots | Power Good Voltage Level out of Limits | Check Power Supply with DMM; Replace Power Supply if Defective |
| Power Supply | Power Supply Fails after Additional Components are Added to System | New Components Require more 5V Power than Old Power Supply can Provide | Replace Failed Unit with a 300-watt or Larger Unit |
| Power Supply | Hard Disk or Fan won't Turn | Defective or Overloaded Power Supply | Replace Failed Unit with a 300-watt or Larger Unit |
| Power Supply | No Leds, No Fans are Turn | Defective Power Supply | Replace Power Supply |
| Processor | System won't Start After New Processor is Installed | Processor not Properly Installed | Reseat and Reinstall Processor and Heatsink |
| Processor | Improper CPU Identification during POST | Old BIOS | Upgrade BIOS |
| Serial ATA | Drives are not Recognized by System | Some Systems have Disabled Serial ATA Ports | Enable Onboard Serial ATA ports |
| Serial ATA | Can't use Onboard Serial Port | Port Might be Disabled in BIOS | Enable Port |
| Serial ATA | Conflict between Onboard Serial Port and other Device | IRQ or I/O port Address Conflicts with other Device | Adjust IRQ or I/O port Address in Use, or Disable Port |

15

| Startup | No Live Screen But System is On | Video Card Problem | Replace your Video Card |
| --- | --- | --- | --- |
| Startup | System Beeps Several Times | Fatal Hardware Errors | Check for Any Defective Hardware |
| Startup | System Can't Find any Hard Drive | Boot Priority Errors | Set Hard Drive as the 1st Booting Device |
| Startup | Computer won't Start After Installing Sound/Video/Network Card | Conflict or Defective Hardware | Remove all Connected Cards and Try Again |

## 3.3. The Inference Engine

It is the brain of the system which performs logical reasoning on rules and problem-solving strategies to derive answers and conclusions, and infers new knowledge. It is fed by troubleshooting data and facts from users and produces results about the causes of the corresponding faults and their possible solutions, keeping the reasoning process totally invisible from the user. The key elements of the inference engine are three: the rules representing troubleshooting knowledge, facts representing circumstances about a certain situation in the real world such as particular hardware or software fault, and the working memory which contains the facts and other runtime parameters to support the inference engine. The job of the inference engine is to match facts in the working memory against rules from the rule-base so as to determine which rule is applicable to which fact and consequently to formulate the conclusion. For instance, the inference engine can match the fact that a "PC cannot start" with the rule that states "IF PC cannot start THEN defective power supply → replace power supply" to conclude that the PC cannot boot because the power supply is defective which should be replaced. The algorithm used behind the inference engine employs a decision tree data structure that models all possible decisions with their possible consequences (Figure 3). The internal nodes of the tree correspond to the troubleshooting questions, the links to user's answer alternatives, and the leaf nodes to conclusions and solutions. As for the reasoning algorithm, the forward-chaining algorithm is used which starts by questioning the user who does know anything about the solution and investigates progressively to reach the diagnosis results and propose some reasonable solutions. Below is the pseudo-code of the forward-chaining algorithm used by the inference engine of the proposed Expert PC Troubleshooter.

Step 1: Read initial facts and store them into working memory.
Step 2: Check the condition part (left side) of every production rule in the rule-base
Step 3: If all the conditions are matched, fire the rule (execute the right side).
Step 4: If more facts are present, do the following:
Step 5: Read next fact and update working memory with the new facts.
Step 6: Go to step 2
Step 7: If more than one rule is selected, use the conflict resolution strategy to select the most appropriate rules and go to step 4.
Step 6: Continue until all facts are exhausted.

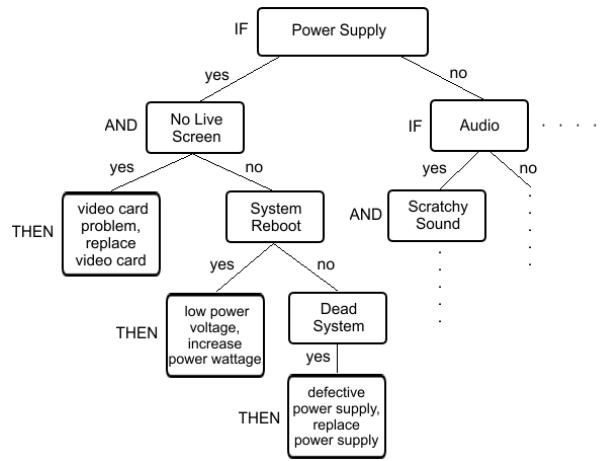

Figure 3. Sample decision tree

## 3.4. The Fuzzy-Logic Inference Engine

The proposed Expert PC Troubleshooter features a fuzzy-logic inference engine and a fuzzy-logic rule-base to diagnose and analyze the BIOS POST beep error codes for the original IBM-PC. POST short for power-on self-test is a set of diagnostic routines built-in the BIOS firmware that check if the computer meets the necessary system requirements and that all hardware and peripherals are working correctly before booting the operating system [13]. If a malfunction is detected, an error message is displayed on the screen and various beep codes are output by the computer's built-in loud speaker. Inherently, fuzzy-logic is a multivalued logic originally proposed by Lotfi Zadeh in 1965 [14] that allows the representation of data values where boundaries are not well defined such as very long, too short, rather cold, extremely hot etc. so that they get processed by machines. Fuzzy-logic is intended to represent a linguistic type of knowledge and reason on this knowledge using a computer. Using fuzzy-logic allows the expression of knowledge in a verbal form making the system more readable, flexible, and user-friendly.

The proposed Expert PC Troubleshooter accepts linguistic data variable such as short, long, and very long representing the duration of the beep generated during the POST process. It then de-fuzzifies it into equivalent crisp (integer) value. Subsequently, the fuzzy-logic inference engine maps it against the rules from the fuzzy-logic rule-base to identify the type of POST fault and finally display the corresponding textual error message on the screen. The fuzzy-logic rules are expressed in the form of *IF variable IS property THEN action*, and they are the following:

IF beep IS very short THEN normal POST, system is OK
IF beep IS short THEN POST error
IF beep IS long THEN system board problem
IF beep IS very long THEN 3270 keyboard card
IF beep IS continuous THEN power supply, system board, or keyboard problem
IF beep IS infinite THEN power supply or system board problem or keyboard

The framework of the fuzzy-logic inference engine can be mathematically defined as ($X$, $Lx$, $\chi$, $\mu x$), where $X$ denotes the linguistic input variable, $Lx$ denotes the different linguistic values that $X$ can take, $\chi$ denotes the set of all possible crisp values, and $\mu x$ denotes a fuzzy function that maps linguistic values into equivalent crisp values [15]. Therefore, the proposed Expert PC Troubleshooter can be described as follows:

$X$ = POST beep duration
$Lx$ = [very short, short, long, very long, continuous, infinite]
$\chi$ = [0, >5]
$\mu x$ is the fuzzy membership function depicted in Figure 4

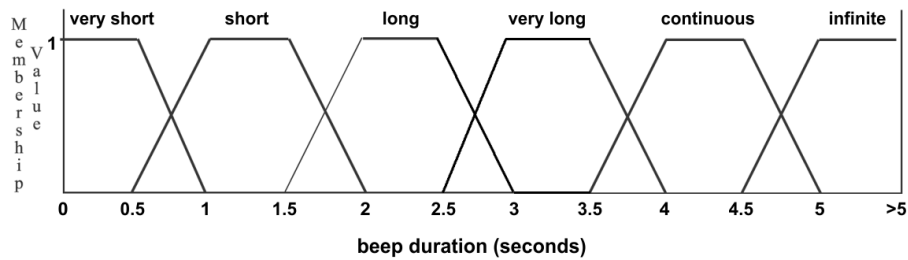

Figure 4. The Membership Function

## 3.5. Intelligent Agent & Knowledge Acquisition

Knowledge acquisition is a method of learning which includes the collection, validation, translation, modeling, and storage of knowledge extracted from various sources such as human experts, other expert systems, or public knowledge repositories [16]. The proposed Expert PC Troubleshooter uses an autonomous intelligent agent whose aim is to acquire knowledge and keep the system's rule-base up to date. By definition an agent is a proactive computational entity that acts on behalf of other entities in an autonomous fashion and exhibits properties like learning, cooperation, and mobility [15]. The function of the proposed agent is to automatically collect knowledge from the World Wide Web by periodically connecting to the Internet and accessing several predefined web pages that contain a wealth of PC troubleshooting rules posted in HTML format. The agent then parses the HTML pages and extracts the embedded rules from within the pages. The agent then checks if these newly scraped rules are already in the knowledge-base, if so, they are ignored; otherwise they are added to the existing rules. As a matter of fact, the intelligent agent is helping in self-learning by automating the knowledge acquisition process which increases the knowledge of the expert system and improves the accuracy of its decision making capabilities.

## 3.6. The Expert Interface

The expert interface is the admin portal of the Expert PC Troubleshooter. It supports the management of production rules stored in the rule-base and allows human experts to review existing knowledge and adding new knowledge to the system. The supported activities are adding new rules, viewing, editing, and deleting existing rules. Figure 5 shows the admin expert interface.

International Journal of Artificial Intelligence & Applications (IJAIA), Vol.3, No.2, March 2012

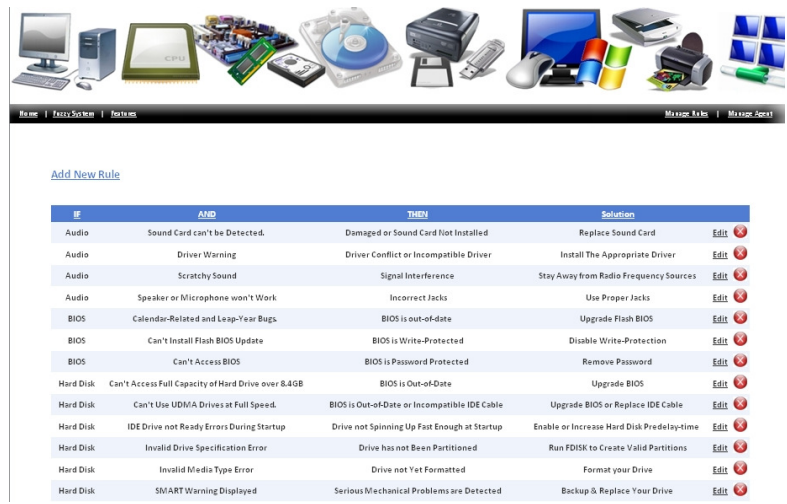

Figure 5. The admin expert interface

## 4. IMPLEMENTATION

The Expert PC Troubleshooter is implemented using ASP.NET 4.0 and C#.NET 4.0 under the MS .NET Framework 4.0 and MS Visual Studio 2010. The rule-base is implemented as a relational database using MS Office Access 2010. Figure 6 shows a source-code snippet of a method that adds new rules to the rule-base via the expert interface. Figure 7 is a snapshot of the rule-base along with a bunch of troubleshooting rules implemented as a database using MS Office Access 2010.

```
protected void okButton_Click(object sender, ImageClickEventArgs e)
{
    string IF = IFTextbox.Text.Trim().Replace("'", "''");
    string AND = ANDTextbox.Text.Trim().Replace("'", "''");
    string THEN = THENTextbox.Text.Trim().Replace("'", "''");
    string solution = solutionTextbox.Text.Trim().Replace("'", "''");

    DatabaseUtilities DatabaseUtilities_OBJ = new DatabaseUtilities();

    DatabaseUtilities_OBJ.ExecuteQuery("INSERT INTO Rules (re_conditionA,re_conditionB,re_conclusion,re_solution)

    addRulePanel.Visible = false;

    GridView1.DataSourceID = "AccessDataSource1"; // Refresh GRIDVIEW
}
```

Figure 6. Source-code snippet

Figure 7. The rule-base in MS Office Access

## 5. CONCLUSIONS & FUTURE WORK

This paper presented a novel expert system for troubleshooting and solving PC problems and faults. The system is called Expert PC Troubleshooter and it supports reasoning on knowledge using an inference engine and a rule-base encompassing troubleshooting production rules, in addition to a user and an admin interfaces that simplify operating and managing the system. Furthermore, the system features a fuzzy-logic module that diagnoses POST beep errors for IBM compatible PCs, and an autonomous intelligent agent that helps in knowledge acquisition. All in all, the Expert PC Troubleshooter allows the diagnosis of faults arising in computers in an accurate, rapid, systematic, efficient, and automated manner relieving human technicians from the burden of regularly scheduled maintenance tasks and routine support issues. Handling POST beep errors in a fuzzy-logic framework using linguistic terms instead of numbers makes the system more readable, easier to understand, flexible, and user-friendly. Likewise, automating the knowledge acquisition process makes the system dynamic and self-learner able to stay up-to-date with the latest troubleshooting knowledge upgrades all the time. In a business context, the Expert PC Troubleshooter meets industrial needs in a cost effective way, reduces maintenance costs, and promotes maintenance, repair, and operations concept (MRO), and automates problem solving and decision making in complex and large organizations and enterprises.

As for future work, the Expert PC Troubleshooter is to be parallelized by implementing a parallel forward chaining algorithm allowing the inference engine to take advantage of multiprocessor systems. In effect, moving to a parallel architecture would increase the overall performance of the expert system and would speed-up its reasoning process.

## ACKNOWLEDGMENTS

This research was funded by the Lebanese Association for Computational Sciences (LACSC), Beirut, Lebanon, under the "Expert System Troubleshooting Research Project – ESTRP2012".

## REFERENCES

[1] Jackson & Peter, (1998) *Introduction to Expert Systems*, 3$^{rd}$ ed, Addison Wesley.
[2] Russell & Norvig, (2003) *Artificial Intelligence: A Modern Approach*, 2$^{nd}$ ed, Upper Saddle River, Prentice Hall.
[3] Lederberg, Joshua, (1987) "How Dendral Was Conceived and Born", *ACM Symposium on the History of Medical Informatics*, Rockefeller University, New York, National Library of Medicine.
[4] Lindsay, Robert K., Buchanan, Bruce G., Feigenbaum, Edward A., & Lederberg, Joshua, (1980) *Applications of Artificial Intelligence for Organic Chemistry: The Dendral Project*, McGraw-Hill.
[5] Shu-Hsien, Liao, (2005) "Expert system methodologies and applications, a decade review from 1995 to 2004", *Expert Systems with Applications*, Vol. 28, No. 1, pp93-103.
[6] Lindsay, Robert K., Buchanan, Bruce G., Feigenbaum, Edward A., & Lederberg, Joshua, (1993) "DENDRAL: A Case Study of the First Expert System for Scientific Hypothesis Formation", *Artificial Intelligence,* Vol. 61, No. 2, pp209-261.
[7] Gupta, Kalyan, (1999) "Case-Based Troubleshooting Knowledge Management", *AAAI Technical Report,* pp99-04.
[8] Verma, Jindal &Aggarwal, Jain, (2010) "An Approach towards designing of Car Troubleshooting Expert System", *International Journal of Computer Applications*, Vol. 1, No. 23, pp107-114.
[9] Yusong, P., Hans, P., Veeke, M. & Lodcwijks, G. (2006) "A Simulation Based Expert System for Process Diagnosis", *In Proceedings of EUROSIS 4th International Industrial Simulation Conference (ISC 2006),* Italy, pp393-398.
[10] Gupta, Kalyan, (1998) "Knowledge-Based Systems for Troubleshooting Complex Equipment", *International Journal of Information and Computing Sciences*, Vol. 1, No. 1, pp. 30-31.
[11] Vaez-Ghaemi, R, (1989) "Knowledge-based troubleshooting and diagnosis of electronic devices", *Instrumentation and Measurement Technology Conference, IMTC-89*, Conference Record, 6th IEEE.
[12] Isa, Sidek, (2000) "PC diagnosis and troubleshooting expert system based on computer response during power-on-self-test (POST) – PCDIASHOOT", *TENCON Proceedings*, Vol. 3, pp458-463.
[13] Mueller, Scott, (2011) *Upgrading and Repairing PCs*, 20$^{th}$ edition, Que publishers.
[14] Zadeh, L.A., (1965) "Fuzzy sets", *Information and Control*, Vol. 8 No. 3 pp338–353.
[15] Akerkar, Rajendra & Sajja, Priti, (2009) *Knowledge-Based Systems*, Jones & Bartlett Publishers.
[16] Raoult, O., (1989) "A Survey of Diagnosis Expert Systems", *in Knowledge-Based Systems for Test and Diagnosis*, Elsevier Science, New York, NY, pp.153-167.